# MISm: A Medical Image Segmentation Metric for Evaluation of weak labeled Data

Dennis HARTMANN[1], Verena SCHMID[1,2], Philip MEYER[2], Iñaki SOTO-REY[2], Dominik MÜLLER[1,2,*] and Frank KRAMER[1]

[1]IT-Infrastructure for Translational Medical Research, University of Augsburg, Germany
[2]Medical Data Integration Center, Institute for Digital Medicine, University Hospital Augsburg, Germany

**Abstract.** Performance measures are an important tool for assessing and comparing different medical image segmentation algorithms. Unfortunately, the current measures have their weaknesses when it comes to assessing certain edge cases. These limitations arouse when images with a very small region of interest or without a region of interest at all are assessed. As a solution for these limitations, we propose a new medical image segmentation metric: MISm. To evaluate MISm, the popular metrics in the medical image segmentation and MISm were compared using images of magnet resonance tomography from several scenarios. In order to allow application in the community and reproducibility of experimental results, we included MISm in the publicly available evaluation framework MISeval: https://github.com/frankkramer-lab/miseval/tree/master/miseval

## 1. Introduction

Machine learning algorithms have become increasingly popular in the field of medical image analysis. Neural networks in particular are a driver of this trend due to their state-of-the-art precision [1–3]. One area of medical image analysis in which, complex problems are investigated, is medical image segmentation (MIS). It is defined as the classification at pixel level of medical imaging data such as computer tomography (CT), magnetic resonance imaging (MRI), or optical coherence tomography (OCT). In this field, aspects such as the localization of organs, vessels, and tumors are examined. Because MIS can be integrated into the decision-making process of clinicians, the quality of the predictions is extremely important. In order to quantify the performance, appropriate evaluation measurements are required. However, it has been demonstrated that the current measurements have limitations in covering certain edge cases like weak labels [4–8]. Weak labels, i.e., data with no area of interest in the segmentation mask, are the cases we want to focus on in this paper. These cases are important for evaluation in the medical field, as control patients are common in clinical trials. However, predictions of weak labels are not taken into account and rated zero by the dice similarity score, regardless of the correctness of the prediction [9]. It is crucial that control patients' imaging data are correctly segmented, which is why a metric covering these edge cases is essential.

---

[*]Corresponding Dominik, Müller, IT Infrastructure for Translational Medical Research,
Alter Postweg 101, 86159 Augsburg, Germany; E-mail: Dominik.Müller@informatik.uni-augsburg.de.

The research aims to develop a MIS metric that covers the limitations of current widely used measurements in evaluating weak labeled data.

## 2. Methods

### 2.1. Metric Definition

Throughout the paper, the definition of a metric from Taha et al. is used [6]. In this paper, we introduce the MISm: **M**edical **I**mage **S**egmentation **m**etric, which is capable of solving the issues with weak labeled data of most metrics in MIS.

$$MISm = \begin{cases} \dfrac{2TP}{2TP + FP + FN} & \text{if } P > 0 \\ \dfrac{\alpha * TN}{(1-\alpha) * FP + \alpha * TN} & \text{if } P = 0 \end{cases} \quad (1)$$

The operators are based on the computation of a confusion matrix for binary segmentation, which contains the number of true positive (TP), false positive (FP), true negative (TN), and false negative (FN) predictions. P is the number of the actual positive conditions and therefore the sum of TP and FN. Furthermore, the weighting coefficient $\alpha \in [0, 1]$ was introduced.
 The first part of our formula is equivalent to the dice similarity or F1-score (DSC). The second part is the Specificity (Spec), also called True Negative Rate, where additional weights α and (1 – α) were added, which was defined as weighted Specificity ($wSpec_\alpha$).
Analogous to the DSC and the Specificity, MISm returns values between [0, 1] with one being equal to the ground truth annotation and zero implying no overlap between ground truth and prediction.

### 2.2. Metric Verification

For the performance measurement of a prediction, common metrics in MIS entail two main limitations. If there is no ground truth annotation, meaning P = 0 and, therefore, TP = FN = 0, then DSC, which is widely used in MIS [2,6,10], takes zero. For FP = 0 or close to zero, the segmentation is an actually accurate true negative, which contradicts the DSC being zero or non-defined. To measure the impact of TN and FP in this case, it is possible to estimate the false positive rate (FPR), also called fall-out, with the formula

$$FPR = \frac{FP}{N} = \frac{FP}{FP + TN} \quad (2)$$

where N is the number of the actual negative condition. If everything is predicted incorrectly, so TN = 0, FPR takes one. To have the common metric range where one means a perfect prediction and zero an incorrect prediction, FPR is reversed and transformed into the Specificity, also called true negative rate, which is commonly used in medicine.

$$1 - FPR = 1 - \frac{FP}{FP + TN} = \frac{FP + TN}{FP + TN} - \frac{FP}{FP + TN} = \frac{TN}{FP + TN} = Spec \quad (3)$$

The application of the Specificity results in the second limitation for the performance measurement of a prediction: Assume N = 60,000 and P = 0. Let FP = 5,000, thus TN = N - FP = 55,000. As almost 10% were predicted false positive, a MIS prediction can be postulated as inaccurate. However,

$$Spec = \frac{55,000}{5,000 + 55,000} \approx 0.9167 \quad (4)$$

implying an acceptable model. To fix this inaccuracy, FP and TN were weighted to each other by adding weights α and 1 – α to the formula.

$$wSpec_\alpha = \frac{\alpha * TN}{(1 - \alpha) * FP + \alpha * TN} \quad (5)$$

In the example above, let α = 0.1. This yields to

$$wSpec_{0.1} = \frac{0.1 * 55,000}{(1 - 0.1) * 5,000 + 0.1 * 55,000} = 0.55 \quad (6)$$

which results in a more insightful scoring and represents the second part of the MISm. The detailed analysis in an experimental setting of the coefficient α can be found in the Results section.

### 2.3. Availability

The MISm has been included in MISeval: an open-source Python framework for the evaluation of predictions in the field of medical image segmentation. This framework is available on Github: https://github.com/frankkramer-lab/miseval
Our code is licensed under the open-source GNU General Public License Version 3 (GPL-3.0 License), which allows free usage and modification for anyone.

## 3. Results

### 3.1. Theoretical Analysis

In the following theoretical analysis, definition gaps, as well as appropriate scoring gradient compared to the DSC and Spec, were investigated. Regarding the definition gap of the Spec

$$TN + FP = 0 \Leftrightarrow TN = FP = 0. \quad (7)$$

The Spec is not defined, if TP ≥ 0, FN ≥ 0, TN = 0 and FP = 0. As P = TP + FN ≥ 0, MISm computes the DSC and is, therefore, defined, although the Spec is not. Analogously, the definition gap of the DSC

$$2TP + FP + FN = 0 \Leftrightarrow TP = FP = FN = 0. \qquad (8)$$

Thus, the DSC is not defined for TN ≥ 0 and the other values being zero, which represents a completely true negative prediction. MISm handles this edge case separately by computing the weighted Specificity as P = 0. Therefore, we conclude that our MISm is always defined.

Analyzing the scoring gradient, we regard P = 0, but let FP > 0. In this case, DSC = 0 and any prediction will yield the same score, despite a strong possible variation of FP. The fixed scoring outcome is not capable of reflecting the prediction quality, properly. In contrast, the Spec grades the predictions, but underweights FP as seen in the example above. As MISm utilizes a weighted Spec in the case P = 0, a more appropriate scoring gradient is sustained.

### 3.2. Weighting Coefficient Analysis

Different weighting coefficients were investigated for our metric and their impact on scoring capabilities on the edge case, in which no predictions are present in the mask. To visualize the score with different weighting coefficients and to compare it with popular metrics of the research field, the different scores in comparison to the ratio of pixels falsely classified as positive to pixels classified as negative, for the edge case (P = 0), were plotted in Fig. 1.

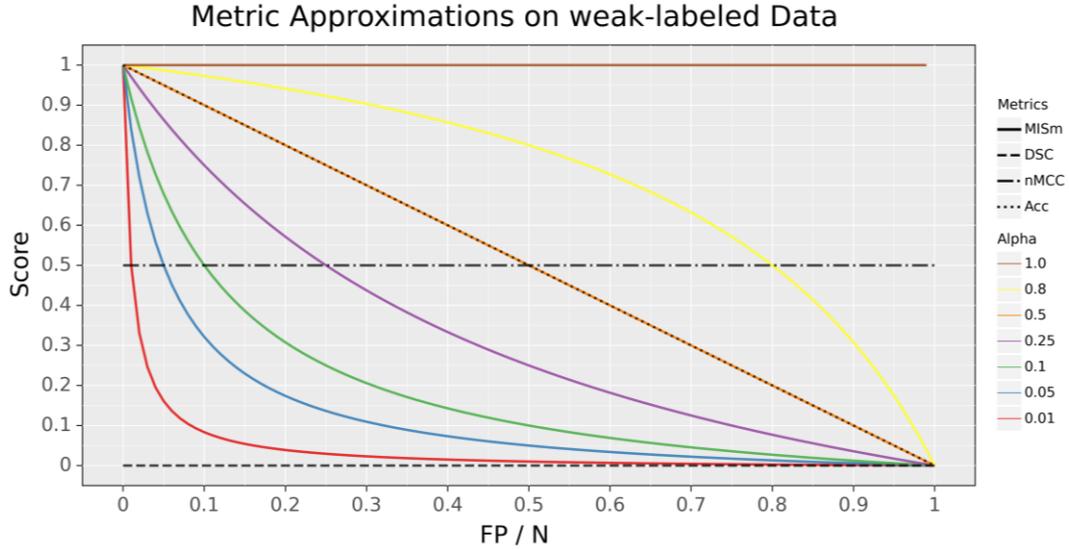

**Fig. 1**: Comparison of the performance metrics considered with our presented metric in terms of the ratio of false positives to actual negatives if the class observed is not present on the image. The experiments with MISm and different weights were also included in the figure.

Even though MISm provides a foundation for the evaluation of datasets with control samples, the weighting factor α is a dynamic variable that results in inconsistent performance assessments by varying weighting factors. The selection of the weighting factor α is still a subjective definition of the assessor, which causes the usage of MISm for quantitative evaluation as ineffective due to the incomparability as consequence. In order to utilize MISm as a generic evaluation method for objective performance measurement, it is mandatory to use a fixed and community-accepted weighting factor. The author proposes the weighting factor as α = 0.1, which is implemented in the software MISeval as the default weight for MISm.

### 3.3. Experimental Application

For experimental application, MISm was compared with popular metrics, such as Accuracy (Acc) [6], Dice Similarity Coefficient (DSC) [6], normalized Matthews Correlation Coefficient (nMCC) [4], and weighted Specificity (wSpec), for MRI brain tumor segmentation, which can be seen in Fig. 2. Part A shows an MRI scan of the brain with an annotated tumor [11,12]. Based on the annotation, various predicted segmentation cases were tested for evaluation (P > 0). In part B, the edge cases in which no tumor is present in the image (P = 0), were illustrated [13]. Popular MIS evaluation metrics and the proposed MISm were calculated for the respective cases to allow comparability between them. For wSpec and MISm, $\alpha$ as 0.1 was selected.

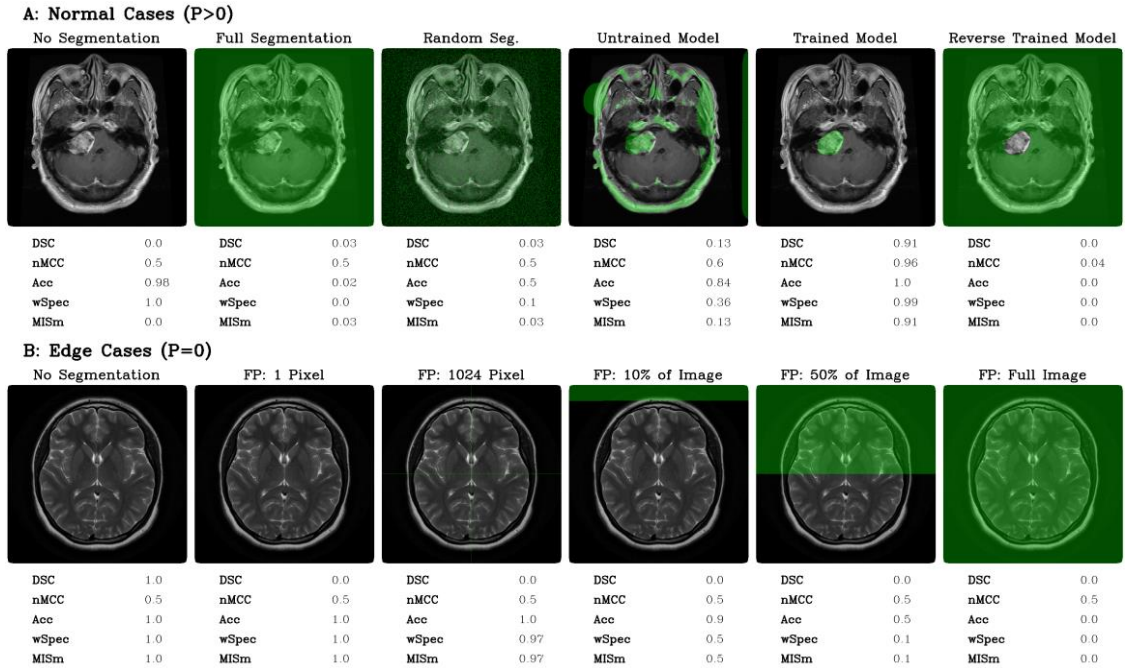

**Fig. 2**: Scoring comparison between MISm and multiple common MIS metrics by application on normal as well as edge cases.

## 4. Discussion

MISm proposes a solution to the limitations identified within current gold-standard metrics popular in the field of MIS. By utilizing wSpec, MISm allows evaluating datasets with weak label annotations. To guarantee flexibility in the usage, the weighting $\alpha$ was designed flexibly. However, for performance assessment in the context of evaluation, it is strongly recommended to use the proposed fixed weighting coefficient α = 0.1. Summarizing, it was proven that MISm is an always applicable metric that is suitable for appropriate prediction scoring.

The MISm equals the DSC if the mask contains an actual positive annotation. As regards, there is no ground truth annotation in the mask, the metrics differ significantly as shown in our theoretical and experimental application. The DSC provides a constant value of zero, whereas the MISm has an adequate scoring gradient by decreasing appropriately for each error, starting at a value of one. In comparison to the nMCC, which is also widely used in several MIS studies [14–17], similar limitations as the DSC were identified, as shown in Fig. **1** and Fig. **2**. Furthermore, interpretation of nMCC is not always intuitive, because a score equal to zero corresponds to an inversed ground truth annotation and 0.5 is equivalent to randomness. All other cases of prediction inaccuracy converge to or take a value of 0.5. This is why the nMCC insufficiently evaluates the quality of the predictions.

The weighting coefficient analysis and experimental application revealed that the Accuracy is capable of scaling in the absence of actual positives. Still, the score is massively influenced by the inclusion of true negatives due to their significantly higher number in MIS. Even so, the Accuracy is capable of handling our identified edge cases, the true negative inclusion constraints the application in practical settings.

### 4.1. Future Work

As future work, the capabilities of the metric MISm as a loss function for model training are quite promising. Currently, the inclusion of control samples in the training process based on an MIS dataset is rare due to the difficulties in performance assessment for the loss function as well as the minimal information gain of control samples for the model. However, the significant difficulties in utilizing MIS models in clinical routine [18–22] indicate that current state-of-the-art models from research are often overfitted on their task [23,24]. Passing images to these models with different medical conditions, from healthy patients or with non-relevant imaging abnormalities like artifacts drastically reduces performance or leads to complete malfunctioning of the prediction. Training models with loss functions capable of also scoring control samples could help reduce the overfitting bias and overcome this challenge. Thus, the integration of the MISm as a loss function into popular neural network frameworks like PyTorch and TensorFlow is planned.

## 5. Conclusions

In this paper, the limitations of several popular MIS metrics were identified. This is why, the novel metric MISm was proposed, which calculates meaningful values for common as well as edge cases arising in MIS. Unlike the gold-standard DSC or the nMCC, it was proven that even weak labels are evaluated meaningfully with MISm. Furthermore, it was shown that MISm is not as sensitive to a large number of negative pixels in contrast to the Accuracy. Summarizing, it was demonstrated that MISm can be applied in all possible cases and has an appropriate scaling.

MISm was added to MISeval, a publicly available Python framework for evaluating medical image segmentation: https://github.com/frankkramer-lab/miseval

## 6. Contributions of the Authors

FK and ISR contributed to the coordination, review, and correction of the manuscript. VS, DM, PM, and DH equally contributed to metric development. VS was also in charge of the verification of our metric and manuscript revision. DM was in charge of the coordination, study design conception, implementation, data analysis, interpretation, and manuscript revision. DH was in charge of manuscript drafting and revision. VS, DM, and DH contributed equally to manuscript drafting.

## 7. Conflict of Interest

None declared.

## 8. Funding

This work is a part of the DIFUTURE as well as the OCTOLAB project funded by the German Ministry of Education and Research (Bundesministerium für Bildung und Forschung, BMBF) grant FKZ01ZZ1804E and 13GW0499E.